\documentclass[runningheads]{llncs}
\usepackage[T2A,T1]{fontenc}
\usepackage[utf8]{inputenc}
\usepackage[russian,english]{babel}

\usepackage{cite}
\usepackage{paralist}
\usepackage{booktabs}
\usepackage{graphicx}
\usepackage{hyperref}
\usepackage{paralist}
\usepackage{lscape}
\usepackage{numprint}
\usepackage{array}
\usepackage{setspace}
\usepackage{longtable}
\usepackage{multicol}
\usepackage[ruled, vlined, linesnumbered]{algorithm2e}
\usepackage{amsmath}
\usepackage{enumerate}
\usepackage{xcolor}

\newcommand{\zzru}[1]{{\foreignlanguage{russian}{#1%
}}}

\begin{document}
\title{Benchmarking Multilabel Topic Classification \\ in the Kyrgyz Language}
\titlerunning{Benchmarking Multilabel Topic Classification in the Kyrgyz Language}
%
\author{Anton Alekseev\inst{1,2,3,4}~\orcidID{0000-0001-6456-3329} \and \\
Sergey Nikolenko\inst{1,2,3}~\orcidID{0000-0001-7787-2251} \and~Gulnara~Kabaeva\inst{4}~\orcidID{0000-0003-3001-7201}}
\authorrunning{A. Alekseev et al.}
%
\institute{
Steklov Mathematical Institute at St.~Petersburg, St. Petersburg, Russia \and 
St. Petersburg State University,  St. Petersburg, Russia \and
Kazan (Volga Region) Federal University, Kazan, Russia \and 
Kyrgyz State Technical University n.~a. I.~Razzakov, Bishkek, Kyrgyzstan}
\maketitle              
\begin{abstract}
Kyrgyz is a very underrepresented language in terms of modern natural language processing resources. In this work, we present a new public benchmark for topic classification in Kyrgyz, introducing a dataset based on collected and annotated data from the news site \textit{24.KG} and presenting several baseline models for news classification in the multilabel setting. We train and evaluate both classical statistical and neural models, reporting the scores, discussing the results, and proposing directions for future work.
\end{abstract}

\section{Introduction}\label{sec:intro}
Kyrgyz is {an agglutinative} Turkic language spoken in several countries, notably China and Tajikistan in addition to Kyrgyzstan; it is by no means an endangered language, and several millions of people call it their mother tongue~\cite{mirzakhalov2021turkic}. However, despite a large amount of linguistic work, including computational linguistics (see Section~\ref{sec:related_work}), it is certainly a \emph{low-resource} language, with a very modest number of tools and datasets available in the open for Kyrgyz language processing\footnote{{For a list of tools, corpora, and other language resources for Turkic languages including Kyrgyz, see e.g. \url{https://github.com/alexeyev/awesome-kyrgyz-nlp} and \url{http://ddi.itu.edu.tr/en/toolsandresources}.}}.  
A recent publication~\cite{mirzakhalov2021turkic}, following the taxonomy proposed in~\cite{joshi2020state}, labels Kyrgyz with the ``Scraping By'' status, defined as follows: ``With some amount of unlabeled data, there is a possibility that they could be in a better position in the 'race' in a matter of years. However, this task will take a solid, organized movement that increases awareness about these languages, and also sparks a strong effort to collect labelled datasets for them, seeing as they have almost none.'' Therefore, we believe that a meaningful effort to construct open manually annotated text collections or other reliable resources for Kyrgyz language processing is in great demand; modern NLP, while shifting towards universal models, is still hard to imagine without at least evaluation data.

Text topic classification is a core task in natural language processing and information retrieval~\cite{iir2008}; it is one of the most popular in practice, with applications in advertising~\cite{zhang2008learning}, news aggregation, and many other industries. Very often, topic categorization is posed as a multilabel classification problem since the same text can touch upon multiple topics~\cite{jesseread2011classifier,MADJAROV20123084,1683770}.

In this work, we present virtually the first labeled dataset for text classification in the Kyrgyz language based on the \emph{24.kg} news portal. Moreover, we propose several baseline models and evaluate their results; in this evaluation, we see that {multilingual} models do help to process Kyrgyz, using even primitive stemming and passing from word $n$-grams to symbol $n$-grams quite expectedly help, and deep learning models that we have considered perform better than the best linear models with virtually no hyperparameter search.
Thus, the contributions of our work are threefold: 
\begin{inparaenum}[(1)]
    \item a novel manually labeled dataset for texts in the Kyrgyz language,
    \item several approaches to multi-label Kyrgyz text classification,
    \item proof of concept for the feasibility of multilingual LLMs for Kyrgyz language processing in supervised tasks.
\end{inparaenum}
The paper is organized as follows: Section~\ref{sec:related_work} discusses related work, Section~\ref{sec:data} introduces our dataset, Section~\ref{sec:models} shows baseline models and experimental setup used in our experiments, Section~\ref{sec:results} discusses experimental results, and Section~\ref{sec:conclusion} concludes the paper.

\section{Related Work}\label{sec:related_work}
\paragraph{Topic classification.}

Text topic classification is one of the oldest and best known tasks in information retrieval and natural language processing~\cite{10.1177/21582440221089963,iir2008}. 
It is straightforwardly defined as a supervised learning (classification) task, often expanding into multilabel classification since longer texts are hard to fit into a single topic~\cite{jesseread2011classifier,MADJAROV20123084,1683770}, a problem that is still attracting attention in the latest deep learning context~\cite{10.1145/3077136.3080834,10.1007/s10489-022-04106-x}.
Many approaches have been developed for datasets of different nature:
\begin{inparaenum}[(i)]
    \item news article datasets such as BBC News~\cite{10.1145/1143844.1143892},
Reuter~\cite{10.5555/1005332.1005345}, 20 Newsgroups~\cite{LANG1995331}, or WMT News Crawl~\cite{Lazaridou2021MindTG};
    \item scientific texts such as arXiv abstracts~\cite{Lazaridou2021MindTG}, patents~\cite{Tang_Jiang_Xia_Pitera_Welser_Chawla_2020}, or clinical texts~\cite{DBLP:journals/corr/abs-1807-07425,10.1145/3436369.3436469,10.1093/jamia/ocz149};
    \item social media posts where topics are usually represented by [hash]tags~\cite{dhingra-etal-2016-tweet2vec}, and more.
\end{inparaenum} 
We note especially prior efforts related to text classification for low-resource languages~\cite{fesseha21,ein-dor-etal-2020-active,DBLP:journals/corr/abs-2005-02068,10.1145/3603168,homskiy2023dn}.

\paragraph{Kyrgyz language processing.}
There already exists a large corpus of linguistic research papers dedicated to various aspects of the Kyrgyz language:
\begin{inparaenum}[(1)]
    \item grammar, syntax and morphology modeling~\cite{washington2012finite,karabaeva2013,yiner2016kyrgyz,bakasova2016,boizou2017kyrgyz,israilova2017morphological,israilova2017algo,sadykov2017model,sadykov2018optim,israilova2019,washington2019free,kochkonvaeva2020modeling,toleush2021development,kasieva2020pos,kasieva2020poster,kasieva2021corpus}, including a recent release of \numprint{780} dependency trees~\cite{benli2023} as part of the Universal Dependencies initiative~\cite{nivre-etal-2017-universal},
    \item text-related statistics and construction of corpora/dictionaries~\cite{sadykov2013manas,baisa2015turkic,momunaliev2016,kasieva2021corpus,kasieva2020pos,kasieva2020poster,arycoglu2021},
    \item computer-aided language learning and other educational systems~\cite{cetin2019assisting,karabaeva2015computer},
    \item machine translation~\cite{kochkonbaeva2016,polat2018machine,washington2019free,tukeyev2020},
    \item lexicons and thesauri~\cite{boizou2017kyrgyz,sharipbay2018comparison},
    \item computational linguistics in general~\cite{problems2013},
\end{inparaenum} and more. 
Kyrgyz also appears in multiple works as a part of multilingual research studies, e.g. on multiway machine translation~\cite{mirzakhalov2021evaluating,mirzakhalov2021large} and even text categorization~\cite{li-etal-2020-low}, although the latter uses a different (Arabic) script than our work.

{News articles represent a traditional and widely used text domain, traditionally a valuable source of data both for information retrieval and natural language processing. News-based datasets have found many applications including such non-traditional ones as relation classification~\cite{rusnachenko-etal-2019-distant}. In this work, we concentrate on the news domain primarily  because to the best of our knowledge, for the Kyrgyz language only fiction, news, and Wikipedia articles are readily available online. Collecting social media content, which may also be useful for numerous NLP tasks~\cite{10.1007/978-3-319-27101-9_7,buraya2017towards,moskvichev2018using,AKKNV16,NKK17,TN18,TN16,N16,KKN16}, is also possible but requires significant extra effort for preprocessing; in particular, preprocessing steps for social media sources would have to include language detection since oftentimes people writing in Kyrgyz also publish posts in Russian and other languages.}

Several research efforts on Kyrgyz open corpora and dictionaries are currently in progress~{(see, e.g.,~\cite{kasieva2020pos,kasieva2021corpus,kasieva2020poster})}, but as of 2023, there are still very few manually annotated datasets useful for Kyrgyz language processing. We hope to start filling this gap with this work.

\section{Dataset}\label{sec:data}
\subsection{Annotation}\label{ssec:data_annotation}

With permission of \textit{24.kg}\footnote{\url{https://24.kg/}} editors, we collected \numprint{23283} news articles in Kyrgyz, dated from May 2017 to October 2022. The portal does not provide any topical tags for articles in Kyrgyz, hence we had to either
    match collected articles with possibly available articles in Russian, which are tagged, or
    annotate them with our own topical categories.
%
The original rubrics used at \textit{24.kg} include:
\begin{inparaenum}[(1)]
    \item \zzru{Власть} (government, politics and law),
    \item \zzru{Общество} (society),
    \item \zzru{Экономика} (economics),
    \item \zzru{Происшествия} (accidents, current events),
    \item \zzru{Агент 024} (current events),
    \item \zzru{Спорт} (sports),
    \item \zzru{Техноблог} (tech),
    \item \zzru{Спецпроекты} (special projects),
    \item \zzru{Кыргызча} (articles in Kyrgyz),
    \item English (articles in English),
    \item \zzru{Бизнес} (business).
\end{inparaenum}
Some of the rubrics are clearly not topical (``English'', ``\zzru{Кыргызча}''), some are multi-topic (``\zzru{Спецпроекты}'', ``\zzru{Агент 024}''), and some other topics also turned out to cover very diverse information. Therefore, we had to introduce our own topical labels.

While some general-purpose taxonomies for content classification do exist and are used, e.g., in advertising, including \textit{dmoz}\footnote{\url{https://www.dmoz-odp.org/}; previously \url{https://www.dmoz.org/}.} and IAB\footnote{\url{https://iabtechlab.com/standards/content-taxonomy/}} taxonomies, the label sets there are too broad for the purpose of news classification. Our preliminary experiments with translated news titles zero-shot classification with IAB Tier~1 tags (in the \textit{label-fully-unseen} setting~\cite{yin2019}) yielded poor prediction quality. However, we still consider this direction very promising from the practical point of view and leave it for future research.

To motivate the introduction of a custom set of labels, we have automatically translated\footnote{Google Translate: \url{https://translate.google.com/?sl=ky&tl=en&op=docs}} titles of Kyrgyz articles into English, randomly sampled \numprint{500} out of \numprint{23284} of them {(a subset small enough to annotate in reasonable time yet hopefully large enough to derive meaningful conclusions regarding the topics)}, and obtained their embeddings via the SentenceBERT model~\cite{reimers-2019-sentence-bert} (\texttt{all-mpnet-base-v2}, {the best-performing\footnote{{As of 24.08.2023: \url{https://www.sbert.net/docs/pretrained_models.html}.}}}~fine-tuned MPNet model~\cite{song2020mpnet}). Then, we have grouped the resulting embeddings using agglomerative clustering (Euclidean distance, Ward linkage~\cite{ward1963hierarchical}, other hyperparameters left at default as provided by \emph{scikit-learn} version $1.0$~\cite{scikit-learn}) into \numprint{100} clusters. Note that the~exact clustering procedure and chosen parameters are not of significant importance here; the main idea is to group texts into hopefully small clusters of very similar titles to speed up annotation and, most importantly, to be able to easily invent topic names that are neither too general nor too specific. Note also that we had to translate the titles and apply the model trained on English language data not for the annotation itself but only because there are no good sentence embeddings models for Kyrgyz with reported quality. Where it was impossible to deduce the topics of the article from the title, we made decisions based on the original Kyrgyz news texts. A sample cluster is presented in Table~\ref{tab:clusters_examples}.

\begin{table}[!t]
    \centering\small
    \begin{tabular}{l|l} \toprule
         \textbf{Title} & \textbf{Proposed labels} \\ \midrule
         The presidential candidate who violated traffic rules paid... &law/crime, politics \\
         Cars of drivers who do not pay fines on time will be... & law/crime	\\
         44 percent of the 108,000 fines imposed for violating traffic... & law/crime	\\
         Party candidate was fined 7,500 soms for holding a concert... & law/crime, politics \\
         Fines for garbage thrown from cars have been increased... & law/crime, ecology \\ 
         \bottomrule
    \end{tabular}
    \caption{Sample cluster (\#16).}
    \label{tab:clusters_examples}
\end{table}

\begin{table}[!t]
    \centering\small\setlength{\tabcolsep}{3pt}
    \begin{tabular}{r|c|c|l}
    \toprule
    \textbf{Class label} & \rotatebox{90}{{\textbf{500-1}}} & \rotatebox{90}{{\textbf{500-2}}} & \textbf{Description} \\ 
    \midrule
    politics & 127 & 174 &   Mentions of politicians and political decisions  \\
    law/crime & 126 &  128 & Judiciary and penitentiary sys., legislature, trials, crime  \\
    foreign affairs & 84 & 91 & Any non-Kyrgyzstan-related news \\
    health & 68 & 63 & Health and medicine-related news (mostly COVID19)\\
    local & 43 & 43 & Traffic rerouting, events scheduling \\
    accidents & 41 & 31 &Disasters, fires, road accidents, etc. \\
    econ/finance & 37 & 50 & Money, import-export and labour-related news \\
    society & 36 & 49  &Local initiatives, protests, other citizen-related news    \\
    culture & 32 & 29 &Cultural events and initiatives, celebrity news\\
    citizens abroad & 17 &  11 & Migration questions and Kyrgyz people abroad\\
    sports & 16 & 15  & Awards, announcements, famous sportspeople mentions    \\
    natural hazards & 13 & 5 & Inconveniences and threats due to natural reasons \\
    development & 12 & 25  & Realty, land use and infrastructural development \\
    religion & 11 & 13  & Religion-related news    \\
    science/tech & 9 & 7 & Everything related to science and technology \\
    border & 8 & 9 & Kyrgyzstan's borders-related conflicts and resolutions \\
    education & 7 & 22 &  News on educational procedures/events/institutions \\
    weather & 6 & 4  &Weather forecasts and reports \\
    ecology & 4 &  4 & Ecological initiatives, laws and reports \\
    natural resources & 2 & 0 & Issues related to natural resources \\
    \bottomrule
    \end{tabular}
    \caption{Topical tags for \textit{24.kg}, total number of tags in the 
    {first two annotation batches}
    and their descriptions. The counts do not sum up to $500$ since this is a multilabel task.}
    \label{tab:custom_labels}
\end{table}

The exploratory annotation task was defined as follows: for each cluster, invent a topic name that best describes most if not all titles and use it as the class label. Then correct the label for titles in the cluster that do not fit the invented topic. If multiple topics apply to some of the titles, add more tags where necessary. After that, we make another pass over all \numprint{500} titles since some of the labels were not ``available'' at the beginning, i.e., a~certain ``general'' label might be added to the label set after some of the texts that would be appropriately labeled with it had already been annotated.
As a result, we obtained a refined list of \numprint{20} labels shown in Table~\ref{tab:custom_labels}.
To validate the label set by comparing label distributions with each other, we have annotated \numprint{500} more English translations of the titles using the same set of labels.
We found that the difference in label count distributions in the two sets of \numprint{500} were relatively small, which showed that our annotation was consistent.
Finally, we have annotated \numprint{500} more texts following the same {procedure}. 
We do not disclose the~exact distribution of the labels in~the~final batch for~the~sake of fairness in possible future competitions: knowing the exact number of texts with a certain label might be used as a test data leak to improve results.

\subsection{Data Description}\label{ssec:data_description}
The dataset consists of \numprint{1500} texts, annotated in three sessions as described above. Since the dataset is relatively small, we split it in only two parts: the first two batches of \numprint{500} (i.e., \textit{training} set has \numprint{1000} texts) and the last batch (i.e. the \textit{test} set has \numprint{500} texts).

\begin{table}[!t]\small
    \centering\setlength{\tabcolsep}{3pt}
    \begin{tabular}{c|r|r|r|r|r|r|r}
        \toprule
        \textbf{~} &  \textbf{\#} & \textbf{\#} & & \textbf{\#} &\textbf{Unique} & & \textbf{Unique} \\
        \textbf{~} &  \textbf{texts} & \textbf{sent.} & \textbf{Sent/text} & \textbf{tokens} &\textbf{tokens} & \textbf{Tok/text} & \textbf{stems}\\ \midrule
        \textbf{Train} &   \numprint{1000} & \numprint{7319} & ${7.32}\pm {5.36}$ & \numprint{107556} & \numprint{18958} & $107.56\pm 74.02$   & \numprint{9872}  \\
        \textbf{Test}  &   \numprint{500}   & \numprint{4025} & $8.05\pm 8.78$ & \numprint{57414} & \numprint{12885} & $114.83\pm 101.15$  & \numprint{6924} \\ \bottomrule
    \end{tabular}
    \caption{Dataset statistics; sentences counted via the {\tt sent\_tokenize} method from NLTK~\cite{bird2009natural}. Per-text values show mean value and standard deviation.}
    \label{tab:dataset}
\end{table}

For further application of models based on the bag-of-ngrams approach, the texts had to be split into tokens and, possibly, stemmed/lemmatized. For tokenization, we used the splitting mechanism provided by the Apertium Project morphological analyzer~\cite{apertium,washington2012finite}; to the best of our knowledge, this is the only open source engine for Kyrgyz morphology.
Similarly, for \textit{word normalization} we used the Apertium-Kir~\cite{washington2012finite} FST's token segmentation; since prefixes are uncommon in Kyrgyz, the first segment was used as the stem.
Overall dataset statistics are presented in Table~\ref{tab:dataset}.

\section{Models and Experimental Setup}\label{sec:models}
The resulting dataset is far too small to be used for training, especially for classical models that heavily depend on frequency estimates of various ratios of tokens and n-grams, e.g., models based on the bag-of-words assumption. However, we can use the dataset in cross-validation to make comparisons across models that perform transfer learning.
Still, we include classical approaches into the benchmark as well, since several works have demonstrated that word/character n-gram baselines are sometimes surprisingly competitive, e.g., in entity linking~\cite{alekseev2022medical,savchenko2020ad}, so they should not be ignored even for a relatively small dataset.

We have used grid search to find the best parameters. Since the training set is small and imbalanced in terms of labels, we used 2-fold validation for hyperparameter search with a stratified split into two subsets preserving the label distribution\footnote{Specifically, we used the \emph{IterativeStratification} algorithm from the \emph{scikit-multilearn} library~\cite{2017arXiv170201460S}.}. 
Below we show the considered values and ranges of hyperparameters in addition to the models themselves.

\subsection{Approaches based on the bag-of-ngrams assumption}\label{ssec:bow}

To provide a classic baseline, we have considered several sparse text representations (essentially bags-of-ngrams) and several corresponding models.

\paragraph{Text preprocessing.} We tested several text representations.
First, we tokenized text (Section~\ref{ssec:data_description}) into unigrams, 1-2-grams, 1-2-3-grams, and 2-3-grams. For frequency cutoffs we retained tokens with maximum document frequency (maxdf) of 40\%, 60\%, 80\%, and 100\% and minimum occurrence (mincount) in 2, 5, or 10 documents. We also set the maximum number of features (maxfeat) equal to \numprint{2000} or \numprint{10000}.
In another set of experiments, we used character n-grams: 2-3-grams, 3-4-grams, and 5-6-grams; maxdf for a character n-gram was set to 40\%, 70\%, or 100\%, mincount was 4, 10, and 15, and maxfeat was \numprint{2000} or \numprint{10000}.
Then, having stemmed the texts as in Section~\ref{ssec:data_description}, we have run experiments with the same ``vectorization'' parameters.

\paragraph{Independent classifiers (``Independent'').} In this set of baselines, we train a separate model for every label, using models that are known to perform well for sparse features: logistic regression (both LBFGS and SGD optimization methods), a linear model with hinge loss (linear SVM), and a linear model with Huber loss (usually preferred for regression tasks). 
In the search for the best model, we treated log-loss, hinge loss, and Huber loss as hyperparameters. Experiments with LBFGS were carried out separately, which is reflected in the results table in Section~\ref{sec:results}. For logistic regression with the LBFGS optimizer (that includes $L_2$-regularization), we have tested the performance with regularization strength $C = \frac{1}{\lambda} \in \{0.7, 0.9, 1.0\}$ and limited the number of iterations to \numprint{1000} or \numprint{10000} steps. For other models, apart from the loss function, we have tested the averaging mechanism (enabling/disabling it), $L_1$, and $L_2$-regularizers, and limited the number of iterations to either \numprint{20000} or \numprint{100000} steps.

\paragraph{Binary classifiers chain (``Chain'').} 
In this approach, the base models (which are the same as in the previous paragraph) make predictions in a sequence; the training task for every label in {a} chain includes predictions for previous labels as features. Apart from the hyperparameters listed in the previous paragraph, we have tried different orderings of the prediction chain.

\paragraph{Multilabel k-Nearest Neighbors.} We have added two models based on k-nearest neighbors to the grid search: 
\begin{inparaenum}[(1)]
\item the model \emph{ML-kNN} introduced in~\cite{zhang2007ml}, which uses Bayesian inference to assign labels to test classes based on the standard kNN output,
\item a binary relevance kNN classifier (\textit{BR-kNN}), a similar method introduced in~\cite{EleftheriosSpyromitros2008} that assigns the labels that have been assigned to at least half of the neighbors.
\end{inparaenum}
Although nearest neighbors classifiers are known to perform poorly for high-dimensional vectors (which bags-of-ngrams are), we added them to the task due to them being multi-label \textit{by design}.
We have tested $k \in \{1, 2, 3, 5, 10\}$ neighbors; for \emph{ML-kNN}, we tried different values of the smoothing parameter, $s \in \{0.1, 0.5, 0.7, 1.0\}$.
As a reliable implementation, we used a combination of models from the \emph{scikit-learn} and \emph{scikit-multilearn} \emph{Python} libraries~\cite{scikit-learn,2017arXiv170201460S}.

\subsection{Neural baseline}

As a modern approach to fine-tuning neural networks, we have intentionally selected the most standard method, which is not necessarily state of the art for other languages.
Among multilingual pretrained large language models, one of the most popular ones is {XLM-RoBERTa} (large)\footnote{Available on HuggingFace: \url{https://huggingface.co/xlm-roberta-large}.}, which is essentially a RoBERTa model~\cite{conneau2020unsupervised} trained on a 2.5TB segment of \emph{CommonCrawl} data containing \numprint{100} languages, including Kyrgyz.
We used {XLM-RoBERTa} in the multilabel classification fine-tuning setting, using a ``classification head'' with two linear feedforward layers with dropout and a binary cross-entropy loss. We have used the same split as before as the train-development split to find the best number of epochs ($14$ out of $15$) based on the Jaccard score metric (see below). We used the AdamW optimizer~\cite{adamw} (the AMSGrad version~\cite{reddi2018convergence}), with weight decay set to $0.01$, learning rate set to $0.00002$, and exponential learning rate scheduling with $\gamma$ coefficient set to $1.0$; $\beta_1 = 0.9$, $\beta_2 = 0.999$, $\epsilon = 10^{-8}$. Batch size was set to $4$ mostly due to the equipment-related constraints. Also note that the pretrained {\tt xlm-roberta-large} checkpoint uses byte-pair encoding (bpe)~\cite{sennrich2016neural} as the tokenizer (provided with the model).

\subsection{Evaluation metrics}

Each prediction is a set of labels represented as a vector of 0s and 1s, where 1 means that the corresponding label has been predicted.
Several metrics from ``regular'' binary and multi-class classification can also be applied for multilabel classification. The counterpart of accuracy here is the fraction of exact matches (``Exact''). The $F_1$ measure (a harmonic mean of precision and recall) can be computed for every sample if we treat each binary vector representation of the label set as binary prediction results and then averaged (``F1-sample'' in Table~\ref{tab:results}). Besides, the $F_1$ measure can be computed for each label and, e.g., micro-averaged (``F1-micro'').
We also used metrics unique to the multilabel setting: share of samples where at least one label is predicted correctly (``@l1'') and the Hamming loss (``Hamm''), i.e., the Hamming distance between binary vectors of labels.
Finally, we report the metric we have used for model selection: the sample-averaged Jaccard similarity computed for each pair of  predicted and ground truth label sets; Jaccard similarity between sets $A$ and $B$ is defined as $\frac{|A \cap B|}{|A \cup B|}$.

\section{Results}\label{sec:results}

\begin{table}[!t]
    \centering\small\setlength{\tabcolsep}{3pt}
    \begin{tabular}{l|r|r|r|r|r|r|r}
        \toprule
         \textbf{Configuration} & \rotatebox{90}{\textbf{JaccCV$\uparrow$}} & \rotatebox{90}{\textbf{@l1$\uparrow$}} & \rotatebox{90}{\textbf{Jaccard$\uparrow$}} & \rotatebox{90}{\textbf{Exact$\uparrow$}} & \rotatebox{90}{\textbf{Hamm$\downarrow$}} & \rotatebox{90}{\textbf{F1-micro$\uparrow$}} & \rotatebox{90}{\textbf{F1-sample$\uparrow$}} \\ \midrule 
         \multicolumn{8}{c}{\textbf{Bag-of-Token-Ngrams}} \\ \midrule
         Independent, LBFGS, 1-gram & .390& .59 & .43 & .29 & .06 & .54 & .48 \\         
         Chain, LBFGS, 1-gram & .405 & .63 & .46 & .31 & .06 &.56 & .51 \\ 
         Independent, SGD, hinge loss, 1-2-gram & .465 & .68 & .47 & .29 & .06  & .56 & .53 \\ 
        Chain, SGD, hinge loss, 1-gram & .474 & .68 & .49 & .32 & .06 &  .56 & .55 \\ 
        ML-kNN, 1 neighbor, 0.1-smoothing, 1-gram & .276 & .45 & .30 & .19 & .10 &  .33 & .35 \\ 
        BRML-kNN, 1 neighbor,1-gram & .276 & .45 & .30 & .19 & .10 & .33 & .35 \\  

         \midrule
         \multicolumn{8}{c}{\textbf{Bag-of-Token-Character-Ngrams}} \\ \midrule
         Independent, LBFGS, 2-3-grams & .491 & .69 & .49 & .33 & .06 & .58 & .55 \\ 
        Chain, LBFGS, 2-3-grams & .494 & .69 & .49 & .33 & .06 & .58 & .55 \\ 
        Independent, SGD, hinge loss, 3-4-grams & .521 & .70 & .46 & .26 & .07 & .55 & .54 \\
        Chain, SGD, hinge loss, 3-4-ngrams & .524 & .71 & .48 & .28 & .07 & .55 & .55 \\ 
        ML-kNN, 1 neighbors, 0.1-smoothing, 2-3-gram & .412 & .65 & .42 & .24 & .08 & .48 & .49 \\ 
        BRML-kNN, 1 neighbor, 2-3-gram & .412 & .65 & .42 & .24 & .08 & .48 & .49 \\
    
         \midrule
         \multicolumn{8}{c}{\textbf{Bag-of-Stem-Ngrams}} \\ \midrule                 
        Independent, LBFGS, 1-gram & .451 & .67 & .50 & .34 & .05 & .59 & .55 \\ 
        Chain, LBFGS, 1-gram & .463 & .68 & .51 & .35 &  .05 & .60 & .56 \\ 
        Independent, SGD, log loss, 1-gram & .514 & .74 & .52 & .33 & .06 & .61 & .59 \\ 
        Chain, SGD, 1-gram & .516 & .74 & .54 & .36 &  .06 & .61 & .60 \\ 
        ML-kNN, 1 neighbor, 0.1-smoothing, 1-gram & .345 & .55 & .36 & .21 &  .09 &  .41 & .42 \\ 
        BRML-kNN, 1 neighbor,1-gram & .345 & .55 & .36 & .21 & .09 &  .41 & .42 \\             
         \midrule
         \multicolumn{8}{c}{\textbf{Bag-of-Stem-Character-Ngrams}} \\ \midrule
        Independent, LBFGS, 2-4-grams & .494  & .71 & .52 & .35 &  .06 &.61 & .58 \\ 
        Chain, LBFGS, 5-6-grams & .490 & .70 & .51 & .35 & .06 & .60 & .57 \\
        Independent, SGD, hinge loss, 3-4-grams & .522 & .70 & .49 & .32 & .06 & .58 & .55 \\ 
        Chain, SGD, hinge loss, 3-4-grams & .524 & .69 & .50 & .33 & .06 & .58 & .56 \\ 
        ML-kNN, 1 neighbor, 0.1-smoothing, 3-4-grams & .425 & .65 & .42 & .25 & .08 & .46 & .49 \\ 
        BRML-kNN, 1 neighbor, 3-4-grams & .425 & .65 & .42 & .25 & .08 & .46 & .49 \\ 
                     
         \midrule
         XLM-RoBERTa (with bpe tokenization) & & \textbf{.88} & \textbf{.66} & \textbf{.46}  & \textbf{.04} & \textbf{.72} & \textbf{.73} \\
         \bottomrule
    \end{tabular}
    \caption{Evaluation results: @l1~--- ``at-least-one'', Hamm~--- Hamming loss, JaccCV~--- mean Jaccard score in cross-validation, $\uparrow$~--- more is better, $\downarrow$~--- less is better.}
    \label{tab:results}
\end{table}

Results of our computational experiments are presented in Table~\ref{tab:results}. It clearly demonstrates that employing the multilingual models for supervised tasks with Kyrgyz text data is feasible, since a fine-tuned \emph{XLM-RoBERTa}-based classifier (without any hyperparameter search) outperforms all other approaches. Note that this result has been far from obvious, since, in our preliminary experiments, fine-tuning another popular model {\tt bert-base-multilingual-cased} (in our case essentially BERT~\cite{devlin2019bert} with an added feedforward layer and dropout) did not bring any meaningful results.
Another interesting observation is that while 
\begin{inparaenum}[(i)]
\item Apertium-Kir does not consider the contexts of words,
\item it is not a lemmatizer, and
\item the selected stemming method is very far from perfect,
\end{inparaenum}
even this kind of text normalization does bring improvements compared to the basic bag-of-ngrams approach. Moving from word ngrams to character ngrams also improves the results in most cases, which one could expect since the Kyrgyz language is morphologically rich.

\section{Conclusion}\label{sec:conclusion}
In this work, we have introduced a new annotated text collection in the Kyrgyz language for multilabel topic classification and evaluated several baseline models\footnote{The dataset, baselines, and evaluation code will be released after a Kyrgyz-language-related competition we plan to hold, {at the following URL: \url{https://github.com/alexeyev/kyrgyz-multi-label-topic-classification}}.}. This is one of the first open datasets for the low-resource Kyrgyz language. As for baselines, we have found that while classical baselines can achieve acceptable results, especially after (even primitive) stemming, a straightforward neural baseline achieves significantly better results even with virtually no hyperparameter search.

In the future, we plan to further improve the current labeling scheme, additionally expanding and validating current annotations; we plan to ask multiple experts to label the texts using models trained on currently presented data to speed up the labeling. Then, we plan to increase the dataset size by annotating more news texts. Afterwards, we plan to hold a competition that should uncover state of the art multilabel classification models for the Kyrgyz language news domain. 

{Also, to enhance the benchmark with an arguably even more fair comparison, we plan to:
\begin{inparaenum}[(1)]
\item translate original texts to~English via \emph{Google Translate} and report the~scores of~the~relevant neural models that employ English LLMs as backbones or the scores of zero-shot classification via prompting state of the art generative models such as, e.g., GPT-4~\cite{openai2023gpt4}; 
\item add the~results of~the~\emph{fastText} supervised classification model trained on~our data to~the~benchmark after~publication;
\item study whether using data from a similar domain in~other Turkic languages can help improve classification quality.
\end{inparaenum}}
In general, we hope that the presented dataset will be able to serve as the basis for these and other experiments and become a starting point for novel NLP research for the Kyrgyz language.

\subsubsection{Acknowledgments}
{This work was supported by the Russian Science Foundation grant \#~23-11-00358. We also thank the anonymous reviewers whose comments have allowed us to improve the paper.}

\bibliographystyle{splncs04}
\bibliography{98_kyrgyz,99_references}

\end{document}